%% file: main.tex
\documentclass[10pt,twocolumn,letterpaper]{article}

\usepackage[pagenumbers]{cvpr} 

\input{preamble}

\def\paperID{***} 
\def\confName{CVPR}
\def\confYear{2026}

\title{Exploring Spatiotemporal Feature Propagation for Video-Level Compressive Spectral Reconstruction: Dataset, Model and Benchmark}

\author{
Lijing Cai,  
Zhan Shi,  
Chenglong Huang, 
Jinyao Wu\\ 
Qiping Li,
Zikang Huo,
Linsen Chen, 
Chongde Zi, 
Xun Cao
}

\begin{document}
\maketitle

\input{sec/0_abstract}    
\input{sec/1_intro}
\input{sec/2_related_work}
\input{sec/3_model}
\input{sec/4_recon}
\input{sec/5_experiment}
\input{sec/6_conclusion}

{
    \small
    \bibliographystyle{unsrtnat}
    \bibliography{main}
}


\end{document}

%% file: preamble.tex
\newcommand{\best}[1]{\textbf{#1}}
\newcommand{\second}[1]{\underline{#1}}

\usepackage[utf8]{inputenc} 
\usepackage[T1]{fontenc}    
\usepackage{url}            
\usepackage{booktabs}       
\usepackage{amsfonts}       
\usepackage{nicefrac}       
\usepackage{microtype}      
\definecolor{MyDarkRed}{rgb}{0.0, 0.2, 0.6}
\usepackage{enumitem}  
\usepackage{graphicx}
\usepackage{array}
\usepackage{tabu}      
\usepackage{multirow}       
\usepackage{multicol}       
\usepackage{multirow}       
\usepackage{float}         
\usepackage{makecell}       
\usepackage{booktabs}  
\usepackage{amsmath} 
\usepackage[table,xcdraw]{xcolor}
\usepackage{pifont}
\usepackage{caption}
\usepackage{marvosym}
\definecolor{cvprblue}{rgb}{0.21,0.49,0.74}
\usepackage[pagebackref,breaklinks,colorlinks,allcolors=cvprblue]{hyperref}

%% file: sec/0_abstract.tex
\begin{abstract}
Recently, Spectral Compressive Imaging (SCI) has achieved remarkable success, unlocking significant potential for dynamic spectral vision. However, existing reconstruction methods, primarily image-based, suffer from two limitations: (i) Encoding process masks spatial-spectral features, leading to uncertainty in reconstructing missing information from single compressed measurements, and (ii) The frame-by-frame reconstruction paradigm fails to ensure temporal consistency, which is crucial in the video perception. To address these challenges, this paper seeks to advance spectral reconstruction from the image level to the video level, leveraging the complementary features and temporal continuity across adjacent frames in dynamic scenes. Initially, we construct the first high-quality dynamic hyperspectral image dataset (DynaSpec), comprising 30 sequences obtained through frame-scanning acquisition. Subsequently, we propose the Propagation-Guided Spectral Video Reconstruction Transformer (PG-SVRT), which employs a spatial-then-temporal attention to effectively reconstruct spectral features from abundant video information, while using a bridged token to reduce computational complexity. Finally, we conduct simulation experiments to assess the performance of four SCI systems, and construct a DD-CASSI prototype for real-world data collection and benchmarking. Extensive experiments demonstrate that PG-SVRT achieves superior performance in reconstruction quality, spectral fidelity, and temporal consistency, while maintaining minimal FLOPs. Project page: \url{https://github.com/nju-cite/DynaSpec}.
\end{abstract}

%% file: sec/1_intro.tex
\section{Introduction}\label{sec_introduction}

Compared to RGB images, Hyperspectral Images (HSIs) offer a unique capability to detect spectral properties between various materials \cite{huang2022spectral}, making them highly promising for classification \cite{fauvel2012advances, wang2023dcn, li2024mambahsi}, detection \cite{he2023object, qin2024dmssn}, tracking \cite{xiong2020material, chen2023spirit, li2023learning, yao2024hyperspectral}, and autonomous driving \cite{basterretxea2021hsi, theisen2024hs3}, etc.\par

Traditional hyperspectral imaging systems typically require scanning along either the spatial or spectral dimension, which limits their applicability in dynamic scenes. To overcome this limitation, Spectral Compressive Imaging (SCI)~\cite{cao2016computational, bacca2023computational} has gained significant attention in recent years. As shown in Fig.~\ref{fig_task}(a), SCI employs spatial-spectral encoding to compress 3D data into a 2D measurement, enabling snapshot acquisition. Reconstruction algorithms then leverage sparsity priors to recover the spatial-spectral information. Despite its merits in bandwidth efficiency and acquisition speed, SCI faces two fundamental limitations, as displayed in Fig.~\ref{fig_task}(b): \textbf{(i)} Mask-based encoding inevitably leads to spatial-spectral information loss, making the reconstruction of occluded content inherently uncertain;  \textbf{(ii)} Frame-by-frame reconstruction is prone to temporal isolation---manifested as poor temporal continuity.

\input{figure_tex/figure1}

Fortunately, in the era of video perception, the relevant information from temporal measurement sequences offers a promising solution. As shown in Fig.~\ref{fig_task}(c), a fixed encoding pattern has the potential to differentially capture complementary features across adjacent frames, 
thereby improving the propagation reconstruction of masked information and enhancing temporal coherence. In light of this, achieving video-level HSIs reconstruction from a sequence of compressed measurements emerges as a compelling yet challenging research problem, primarily due to two key obstacles:\\
\textbf{(i)~Data scarcity is a fundamental bottleneck.} Existing datasets are primarily collected for image-level reconstruction \cite{yasuma2010generalized, DeepCASSI:SIGA:2017}. While slicing images to generate pseudo-sequences is an optional workaround \cite{shi2023compact}, such data fails to exhibit high degrees of freedom in motion. In addition, some video-level datasets developed for downstream tasks \cite{xiong2020material} suffer from limited spectral resolution and data reliability, rendering them unsuitable as ground truth for reconstruction-oriented tasks. \textbf{(ii)~Existing algorithms have limited capacity for spatiotemporal modeling.} HSIs reconstruction methods can be broadly categorized into model-based~\cite{beck2009fast, bioucas2007new, yuan2016generalized, liu2018rank} and learning-based approaches~\cite{huang2022spectral, meng2020end, miao2019net, ma2019deep}. Model-based methods require  extensive parameter tuning, making it difficult to achieve stable and high-quality reconstruction. Learning-based methods have recently achieved SOTA performance; however, their high computational cost and limited ability to capture spatiotemporal dependencies pose significant obstacles to video-level reconstruction.

As a first step toward addressing these gaps, we construct a high-quality dynamic hyperspectral image dataset, named DynaSpec. Each frame is captured individually using a push-broom hyperspectral camera, covering the 400--700 nm spectral range with a spectral resolution of 2 nm. Diverse motions are then manually introduced to emulate the high degrees of freedom encountered in real-world scenarios.  The dataset comprises 30 video sequences (totaling  $300~\text{HSIs}$), facilitating the exploration of video-level reconstruction and downstream tasks.

Furthermore, we propose a video-level compressive spectral reconstruction algorithm, PG-SVRT, which consists of three key components:  Mask-Guided Degradation Perception (MGDP), Cross-Domain Propagated Attention (CDPA), and Multi-Domain Feed-Forward Network (MDFFN). MGDP models the degradation process to aid in decoupling intra-frame encoded information. CDPA facilitates progressive cross-domain feature propagation via spatial-then-temporal attention.  Inspired by linear attention~\cite{katharopoulos2020transformers, han2023flatten, han2024agent, han2024bridging}, we introduce bridged tokens to reduce computational complexity while maintaining  high-quality spatiotemporal feature extraction. Additionally, MDFFN allows for the independent extraction of spatial and temporal features while promoting their effective integration.

Finally, we conduct comparative simulation experiments across four SCI systems~\cite{wagadarikar2008single, gehm2007single, cao2011prism, chen2023notch}. The results show that DD-CASSI, benefiting from its high spectral sampling efficiency and clear structural representation, exhibits significant superiority in video-level reconstruction.  Building on this insight, we construct a prototype to capture real-world measurements for validation and benchmarking.

\textbf{In summary, our contributions are as follows:}

\begin{itemize}[leftmargin=1.5em,
                labelsep=0.5em,
                itemsep=0pt,
                topsep=0pt,
                parsep=0pt,
                partopsep=0pt]
  \item We construct the DynaSpec dataset to address the scarcity of high-quality dynamic HSIs data.
  \item We propose a novel method, PG-SVRT, for efficient video-level compressive spectral reconstruction, which achieves over 41 dB PSNR and minimal computational cost, without additional hardware modifications.
  \item We conduct comparative simulations across representative SCI systems, and construct a lab prototype for real-world imaging. Simulations and experiments demonstrate that our method achieves superior performance.
\end{itemize}

%% file: figure_tex/figure1.tex
\begin{figure*}[t!]
\centering
\includegraphics[width=0.93\textwidth]{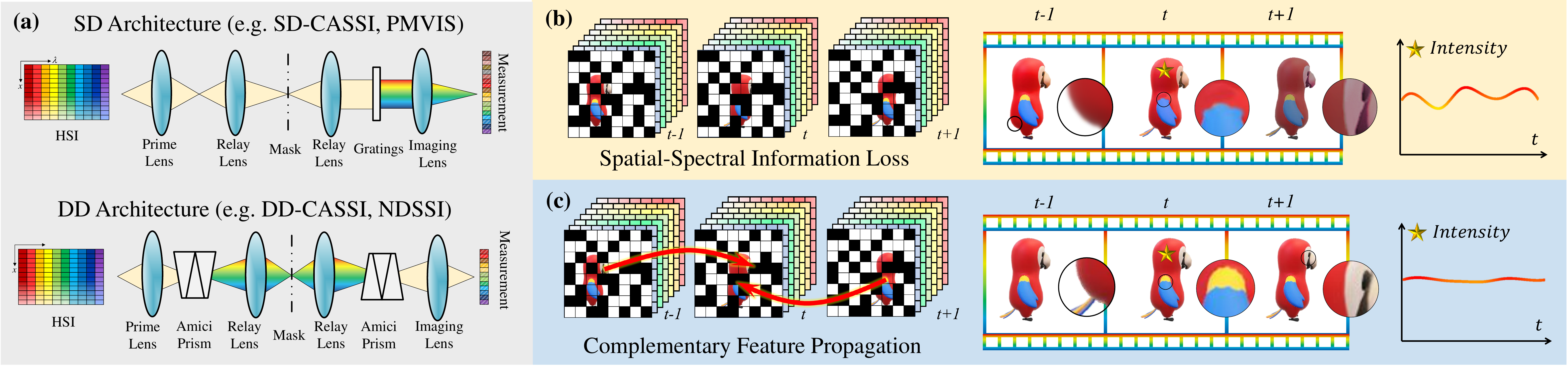}
\caption{Spectral compressive imaging and reconstruction.
(a) SCI principle.
(b) Image-based methods, with issues of uncertain reconstruction and temporal inconsistency (flickering intensity curves).
(c) Video-based reconstruction, where information complementarity enhances completeness and temporal consistency (smooth intensity curves).}
\label{fig_task}
\end{figure*}

%% file: sec/2_related_work.tex
\vspace{-2pt}
\section{Related Work}
\vspace{-3pt}
\textbf{Compressive Spectral Reconstruction.} Model-based approaches typically rely on handcrafted priors such as sparsity~\cite{tan2015compressive}, low-rankness~\cite{liu2018rank}, and total variation~\cite{bioucas2007new, yuan2016generalized}, which struggled with high-dimensional data and complex degradations. Recently, learning-based methods have emerged, including end-to-end models~\cite{cai2022coarse, cai2022mask, xu2023degradation, wang2025s2, hu2022hdnet}, plug-and-play (PnP) methods~\cite{yuan2020plug, yuan2021plug, qiu2021effective}, and deep unfolding networks (DUNs)~\cite{cai2022degradation, meng2023deep, zhang2024dual, li2023pixel, dong2023residual}. PnP methods embed pre-trained denoisers into the optimization process, but the lack of joint training limits their flexibility~\cite{FY_LDP}. DUNs offers interpretability by mimicking iterative solvers with learnable modules, yet suffers from high computational cost~\cite{wang2025s2}.
These methods only consider the reconstruction of single-frame measurements, while video-level spectral reconstruction remains largely unexplored. In this work, we extend the learning paradigm to the video domain, enabling efficient modeling of spatiotemporal dependencies.

\input{figure_tex/figure2}

\textbf{Spectral Compressive Imaging.} SD-CASSI~\cite{wagadarikar2008single} and DD-CASSI~\cite{gehm2007single} employ random binary masks that satisfy compressed sensing requirements but suffer from complicated reconstruction due to severe spectral aliasing.~\cite{cao2016computational}.  PMVIS~\cite{cao2011prism} employs spatially sparse sampling to suppress interference, thereby reducing the ill-posedness of inversion, albeit at the cost of spatial resolution and light throughput. In contrast, NDSSI~\cite{chen2023notch} uses a notch-coded mask to maximize optical throughput and incorporates dual-disperser architecture~\cite{willett2013sparsity} to maintain the fidelity of spatial structures, although its spectral sampling density remains limited. Each system has its strengths and limitations; therefore, we conduct simulation-based evaluations to compare their performance and identify the architecture that delivers superior results for video-level spectral reconstruction.


%% file: figure_tex/figure2.tex
\begin{figure*}[t]
    \centering    \includegraphics[width=0.90\linewidth]{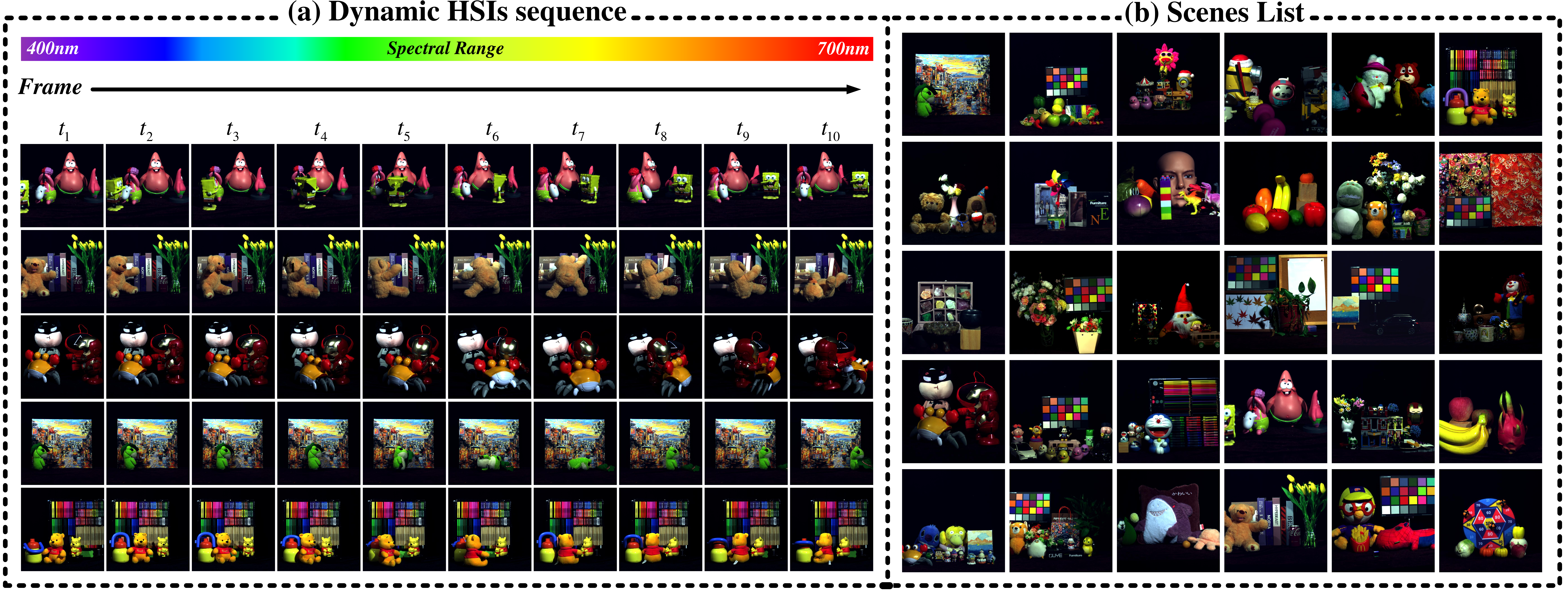}
    \vspace{-8pt}
    \caption{ The proposed DynaSpec dataset. (a) Dynamic HSIs sequences acquired frame by frame to simulate the diverse motion of real-world scenarios. (b) A display of the 30 scenes.}
    \vspace{-8pt}
    \label{fig_dataset}
\end{figure*}

%% file: sec/3_model.tex
\section{Spectral Compressive Imaging Model}
\label{sec_imagemodel}

In this section, we present a unified mathematical formulation of compressive spectral imaging.

\textbf{\textbullet\ Single-Disperser Architecture (SD).}   
As illustrated in the top portion of Fig.~\ref{fig_task}(a), SD systems such as SD-CASSI and PMVIS first modulate the incident light using a coded mask, followed by spectral shearing through a dispersive element. Let $X_{i} \in \mathbb{R}^{H \times W \times C}$ denote the transient 3D spectral data, where $H$ and $W$ are spatial dimensions and $C$ is the number of spectral channels. The measurement at spatial location $(h, w)$ and time $i$ is modeled as:

\vspace*{-18pt}
\begin{equation}
    Y_i(h, w) = \sum_{c=1}^{C} \Phi(h, w) \cdot X_i(h, w - \sigma(c), c)
\end{equation}
\vspace*{-12pt}

where $c$ indexes the spectral coordinate, $\Phi$ denotes the mask modulation function, and $\sigma(\cdot)$ represents the dispersion function introduced by the disperser.

\textbf{\textbullet\ Dual-Disperser Architecture (DD).}  
As shown in the lower part of Fig.~\ref{fig_task}(a), in dual-disperser systems  (e.g., DD-CASSI and NDSSI), the incoming light is first dispersed, then modulated by a coded mask in the spectrally sheared domain, and finally recombined by a symmetric disperser to reverse the spectral shift. We model the measurement at spatial location $(h, w)$ and time $i$ as follows:

\vspace*{-15pt}
\begin{equation}
    Y_i(h, w) = \sum_{c=1}^{C} \Phi(h, w - \sigma(c)) \cdot X_i(h, w, c)
\end{equation}
\vspace*{-12pt}

For convenience, the imaging processes of both architectures are unified and formulated as:

\vspace*{-8pt}
\begin{equation}
    Y_i = \Psi X_i + \Theta
\end{equation}
\vspace*{-12pt}

where $Y_i \in \mathbb{R}^{H \times W'}$ denotes the measurement, $\Psi: \mathbb{R}^{H \times W \times C} \rightarrow \mathbb{R}^{H \times W'}$ denotes the overall encoding operator (e.g., spectral dispersion and mask modulation), and $\Theta$ represents noise. In SD systems, due to spectral shearing, the measurement width is $W' = W + \sigma(C{-}1)$. In contrast, DD systems perform symmetric inverse dispersion, restoring the spatial dimension such that $W' = W$.

Since both architectures can be reformulated under a unified framework, prior SD-based reconstruction methods remain valid and applicable in the context of DD systems. Building upon this, our work is devoted to extending image-level reconstruction to the video level, aiming to recover the spectral video $X \in \mathbb{R}^{ T \times H \times W \times C} $ from the compressed measurement sequence $Y \in \mathbb{R}^{ T \times H \times W'}$.

\vspace{-6pt}
\section{DynaSpec dataset}
\label{sec_dataset}
\vspace{-4pt}

Current datasets for HSIs reconstruction are mainly image-based~\cite{yasuma2010generalized, DeepCASSI:SIGA:2017}. Although pseudo-video sequences can be synthesized by cropping images~\cite{shi2023compact}, such methods only mimic rigid motions, such as camera motion, leaving real-world in-scene dynamics unaccounted for. Some downstream tasks have started exploring spectral data at the video level~\cite{xiong2020material, basterretxea2021hsi, you2019hyperspectral};  however, due to task-specific considerations, the datasets used in these works often suffer from limited spectral resolution and reduced data fidelity, making them unreliable as ground truth for reconstruction tasks. Given this, it is imperative to construct a dataset consisting of high-quality spectral sequences featuring dynamic scenes.

\input{figure_tex/figure3}

Considering the inherent challenges of acquiring ground truth HSIs towards dynamic scenes, we employ the GaiaField push-broom hyperspectral camera~\cite{dualix_gaiafield} to capture controllable objects frame-by-frame. By manually designing actions, we simulate diverse and complex motions, such as translation, rotation, and articulated movements, as shown in Fig.~\ref{fig_dataset}(a). To ensure the authenticity and reliability of the acquired data, we adhere to the following principles: (1) Object motion between consecutive frames is continuous and adheres to physical laws. (2) Long integration times are used  to mitigate noise interference. (3)  Spectral correction is applied based on the camera’s spectral response. (4) The spectral properties of the illumination are excluded to ensure the data approximates reflectance values,  preventing the network from fitting illumination-specific information. 
(5) Intensity calibration is performed using invariant objects within the sequence, minimizing the impact of temperature drift introduced by prolonged system operation.

As illustrated in Fig.~\ref{fig_dataset}(b), the DynaSpec dataset encompasses 30 scenes, totaling 300 HSIs. Each frame’s data cube features a spatial resolution of $1280\times1280$ pixels, a spectral resolution of $2~nm$, and a wavelength range from $400~nm$ to $700~nm$ (151 spectral channels). The dataset’s continuity of motion, coupled with its high spectral and spatial resolution, provides a robust foundation for the reconstruction task.

%% file: figure_tex/figure3.tex
\begin{figure*}[t!]
\centering
\includegraphics[width=0.90\textwidth]{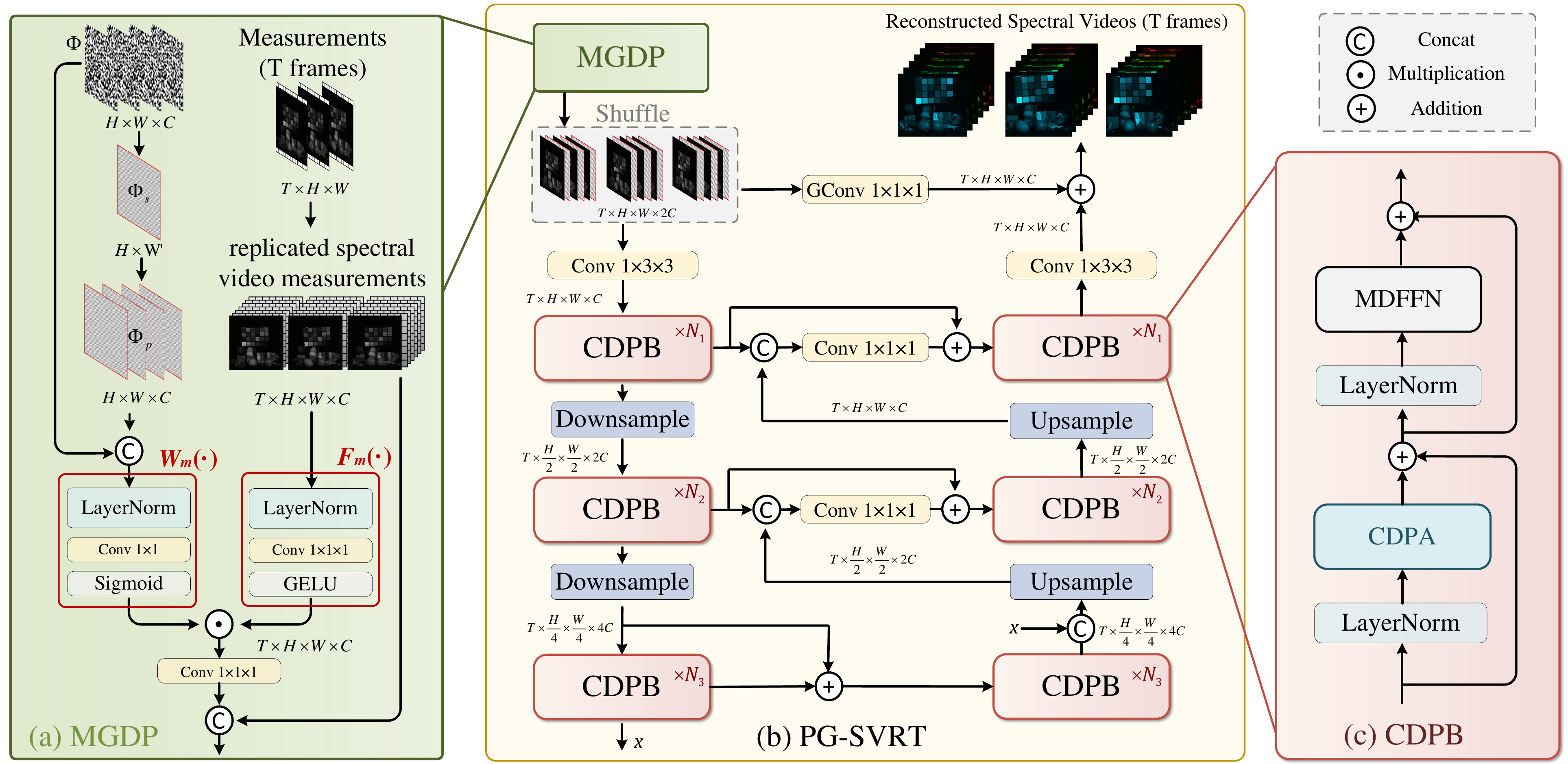}
\vspace*{-0.5em}
\caption{Illustration of PG-SVRT. (a) and (c) The components of MGDP and CDBP. (b) PG-SVRT framework and key components.}
\label{fig_PGSVRT}
\vspace*{-13pt}
\end{figure*}


%% file: sec/4_recon.tex
\vspace{-4pt}
\section{Propagation-Guided Spectral Video Reconstruction Transformer (PG-SVRT)}
\label{sec_pgsvrt}
\vspace{-4pt}

To efficiently extract redundant spatiotemporal features for video-level spectral reconstruction,  we present PG-SVRT, as illustrated in Fig.~\ref{fig_PGSVRT}(b). It utilizes the U-Net-based architecture, primarily composed of the MGDP module and the Cross-Domain Propagated Block (CDPB). The shuffle operation aligns degradation features with measurements across the spectral dimension. The CDPB consists of the CDPA and MDFFN, as shown in Fig.~\ref{fig_PGSVRT}.

\subsection{Mask-Guide Degradation Perception (MGDP)}

Since all measurement sequences follow the same optical encoding paradigm, the degradation prior is essential for reconstruction.  Inspired by the widespread use of mask-based degradation learning in SOTA methods~\cite{cai2022mask,wang2025s2, zhang2024dual,dong2023residual}, we construct the MGDP to perceive the compression degradation process before feeding it into the main architecture. 

The mask matrix $\Phi \in \mathbb{R}^{H \times W \times C}$ in this section represents the mask pattern corresponding to the same spatial region across all spectral channels.  According to Sec. \ref{sec_imagemodel}, for the SD architecture, $\Phi$ for each channel is $\Phi(m,n)$; for the DD architecture, the mask for each channel is given by $\Phi(m,n - \sigma(c))$.   As shown in Fig.~\ref{fig_PGSVRT}(a), we first compress $\Phi$ along the spectral dimension  according to the SD or DD architecture to obtain $\Phi_s \in \mathbb{R}^{H \times W'} $, and then crop or replicate  $\Phi_s$ to form the $\Phi_p \in \mathbb{R}^{H \times W \times C}$. $\Phi_p$ represents the spatial intensity distribution of each channel after degradation. The intensity distribution difference between $\Phi$ and $\Phi_p$ is learned through a $Conv_{1\times1}$, with the sigmoid function used to compute the weight distribution $W_\Phi$, which perceives the degradation features at different spectral and spatial positions. Similar operations are then applied to the measurement sequence $Y$, and element-wise feature weighting is performed by dot-multiplying with $W_\Phi$. Finally, a $Conv_{1\times1\times1}$ convolution is used to extract the mask-guided degradation perception features, which are then concatenated  with the measurements along the channels and serve as the input to the main architecture, denoted as $Y_{in}$, expressed as:

\vspace*{-14pt}
\begin{equation}
    Y_{in} = Concat(Conv(W_{m}(\Phi,\Phi_{p}) \odot F_{m}(Y)),Y)
\end{equation}
\vspace*{-19pt}

\input{figure_tex/figure4}

\subsection{Cross-Domain Propagated Attention (CDPA)}

Spatiotemporal feature extraction~\cite{selva2023video, cai2024exploring} is commonly used in video-level tasks. However, our task also demands the reconstruction of high-dimensional spectral data, posing a critical challenge in designing low-complexity attention mechanisms to ensure computational efficiency.  We consider the following two aspects: (1) Mainstream video attention mechanisms, such as G-MSA~\cite{9741335} and F-MSA~\cite{bertasius2021space}, suffer from high computational complexity, while W-MSA~\cite{liang2024vrt} can reduce the complexity to approximately linear  without compromising feature extraction capability.  However, spectral reconstruction task involves high-dimensional features ($C$), and overly small spatial windows~($H_{win}W_{win}$) are unable to fully capture the dispersion features, creating a bottleneck in complexity reduction. (2) For multi-dimensional feature processing, joint extraction and separated extraction are the two main approaches~\cite{bertasius2021space}. Joint extraction is resource-intensive~\cite{bertasius2021space},   while fully discrete processing of different dimensions  will restrict  the ability for feature interaction.

Given this, we design the CDPA, as shown in Fig.~\ref{fig_CDPB}(a). CDPA is a spatial-then-temporal progressive attention that uses a shared value to propagate features across different domains, alleviating the issue of discrete multi-domain feature interactions. Additionally, inspired by linear attention mechanisms~\cite{han2024bridging}, we introduce a bridged token to further reduce computational complexity.

To simplify the description, we use the CDPA from the first CDPB as an example. The query $ Q_{N1}$, key $K_{N1}$, and value $V_{N1}$ are computed from the input feature $Y_{N1} \in \mathbb{R}^{T \times H \times W \times C}$ as follows:

\vspace*{-15pt}
\begin{equation}
\small
Q_{N1} = Y_{N1} W_q, \quad K_{N1} = Y_{N1} W_k, \quad V_{N1} = Y_{N1} W_v,
\end{equation}
\vspace*{-12pt}

where \( Q_{N1}, K_{N1}, V_{N1} \in \mathbb{R}^{T \times H \times W \times C} \), and \( W_{q,k,v} \in \mathbb{R}^{C \times C} \) are learnable projection matrices.

In spatial information processing, we divide the input into non-overlapping blocks \( Q_s, K_s, V_s \in \mathbb{R}^{T h w \times H_{win} W_{win} \times C} \), where \( h = {H}/{H_{win}}, w = {H}/{W_{win}} \), and \( H_{win}\), \(W_{win}\) denote the window size. Additionally, we introduce a bridged token \( B_s \in \mathbb{R}^{T h w  \times N_B \times C} \), where \( N_B \) denotes the number of tokens. This token serves as a bridge, enabling indirect interaction between  \( Q_s \), \( K_s \), and \( V_s \). Although many advanced methods can effectively generate new tokens~\cite{xia2022vision,bolya2022tome}, since \( B_s \) essentially represents the information of \( Q_s \), we directly pool \( Q_s \)  to generate \( B_s \), avoiding additional computational cost. The final spatial attention can be expressed as:

\vspace*{-15pt}
\begin{equation}
\small
Y_{s}^{out} = GConv \left( A(Q_s, B_s, A(B_s, K_s, V_s, \tau_1), \tau_2)\right) + Y_{N1} 
\end{equation}
\vspace*{-15pt}

where \( A(Q, K, V, \tau) \) represents \( Softmax \left( {Q K^T}/{\tau} \right) V \), with \( \tau \) being learnable parameters~\cite{zhang2024dual}.

In the temporal information processing, we rearrange \( Q_{N1}, K_{N1} \), and \( Y_{s}^{out} \) into \( Q_t, K_t, Y_t \in \mathbb{R}^{H W \times T \times C} \) for temporal attention computation, which not only reduces the resource consumption caused by additional projections but also  enhances feature propagation across different domain through the shared value, directly derived from \( Y_{s}^{out} \).  This computation process is expressed as:

\vspace*{-5pt}
\begin{equation}
Y_{t}^{out} = A(Q_t, K_t, Y_t, \tau_3)
\end{equation}
\vspace*{-12pt}

Based on the above process, the computational complexity of CDPA is displayed as follows:

\vspace*{-12pt}
\begin{equation}
\small
O(CDPA) = 4 T H W C^2 + 4 T H W N_B C + 2 T^2 H W C
\label{eq_complex}
\end{equation}
\vspace*{-14pt}

When \( 2N_B < H_{win}W_{win} \), the introduction of the bridged token reduces computational resource consumption without affecting reconstruction performance. We adopt a rectangle-window strategy~\cite{chen2022cross} with \( H_{win} = 8 \) and \( W_{win} = 32 \), and the bridged token number \( N_B \) is set to 64, satisfying the aforementioned conditions. It is worth noting that we do not set a temporal window. The inherent property of the task suggests that measurements  exhibit strong inter-frame correlations only within short time windows, implying  that $T$ is typically small.  Moreover, temporal windowing may introduce additional padding, which could potentially lead to signal interference or redundant computations.

\subsection{\normalsize\bfseries Multi-Domain Feed-Forward Network (MDFFN)}

Conventional FFNs use a fully connected MLP~\cite{liang2024vrt} or convolution~\cite{cai2022mask} to process Attention outputs, which suffers from high computational complexity and struggles to integrate multi-domain information. Inspired by the multi-head mechanism and 3DCNN decomposition~\cite{tran2018closer}, we propose MDFFN, as shown in Fig.~\ref{fig_CDPB}(b), which divides spectral features into different heads to perform self-attention in both spatial and temporal domains, effectively enhancing in-domain feature extraction capabilities.

%% file: figure_tex/figure4.tex
\begin{figure*}[t!]
\centering
\includegraphics[width=0.95\textwidth]{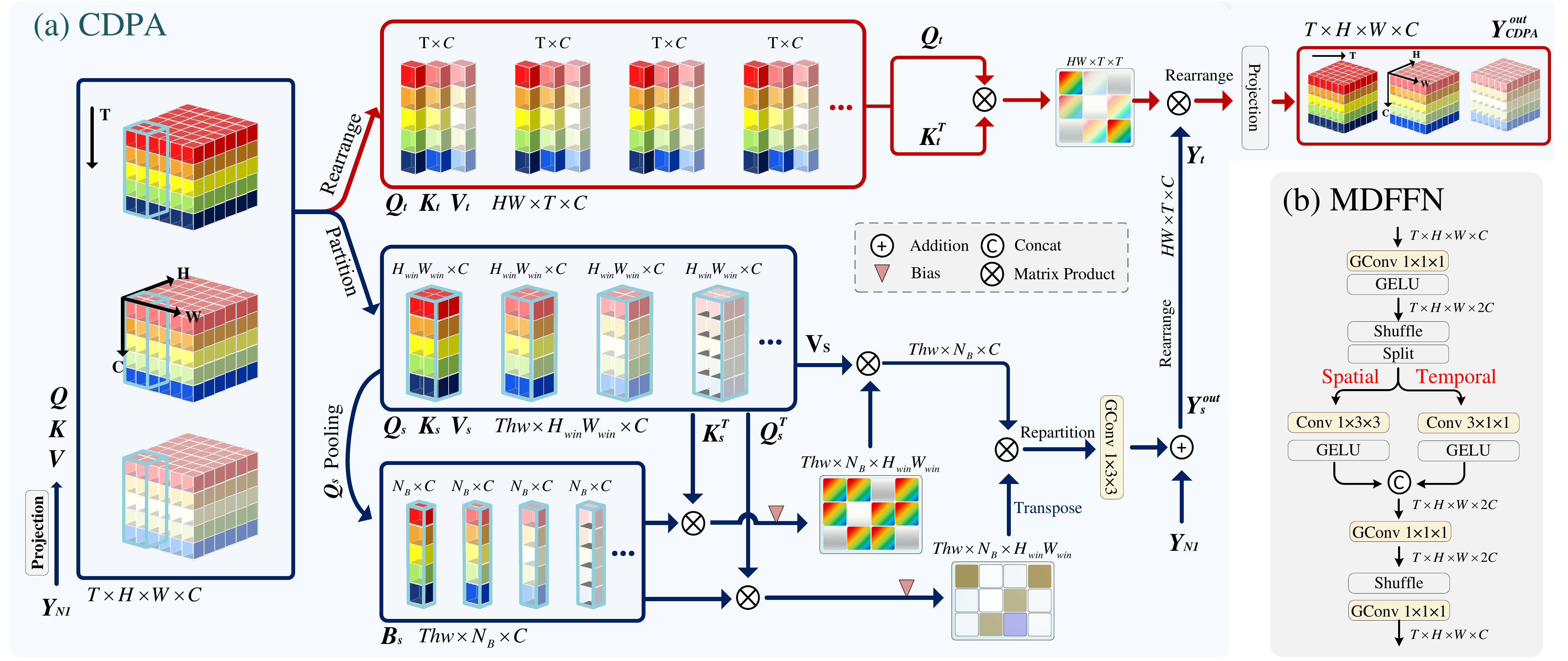}
\vspace*{-0.5em}
\caption{Details of the CDPB, which consists primarily of CDPA and MDFFN. (a) CDPA is a spatial-then-temporal attention mechanism, where the blue line represents spatial feature processing and the red line indicates temporal feature processing. (b) Illustration of MDFFN.}
\label{fig_CDPB}
\vspace*{-13pt}
\end{figure*}

%% file: sec/5_experiment.tex
\vspace{-6pt}
\section{Experiment}
\label{sec_exp}

We initiate our investigation by evaluating representative SCI architectures to compare their potential for video-level spectral reconstruction. Based on the selected system, we further conduct both quantitative and qualitative comparisons to evaluate the reconstruction performance against SOTA algorithms. All experimental bands align with the real system prototype, covering 30 spectral channels from 500 nm to 650 nm, with detailed information provided in the \colorbox{gray!15}{\textcolor{MyDarkRed}{Supp. 1}}.

\subsection{Datasets and Details}
\label{sec_exp_setting}

\textbf{Datasets.} In the simulation, we use three datasets: CAVE~\cite{yasuma2010generalized}, KAIST~\cite{DeepCASSI:SIGA:2017}, and DynaSpec. Referring to previous studies~\cite{cai2022mask}, CAVE and 25 DynaSpec sequences are used for training, and KAIST and the remaining DynaSpec samples serve as the test set. Additionally, the cropping strategy~\cite{shi2023compact} is introduced to generate videos with a spatial resolution of $256\times256$. During training, CAVE is cropped using random steps, while DynaSpec has a 70\% probability that the step size is set to 0, due to its inherent dynamic information. For testing, KAIST and DynaSpec are cropped using pre-randomly generated step sizes. In real experiments, we test five real  measurement sequences captured by the DD-CASSI system, with a spatial size of $1024\times1024$.

\textbf{Comparison Methods.} Due to the lack of research on video-level reconstruction,  we compare PG-SVRT with image-based SOTA methods, including MST-L~\cite{cai2022mask}, CST-L~\cite{cai2022coarse}, DAUHST~\cite{cai2022degradation}, GAP-Net~\cite{meng2023deep}, DADF-Plus-3~\cite{xu2023degradation}, PADUT~\cite{li2023pixel}, RDLUF~\cite{dong2023residual}, $S^2$-Transformer~\cite{wang2025s2},  SSR~\cite{zhang2024improving}, and  DPU~\cite{zhang2024dual}. Additionally, a comparison with RGB video restoration methods TempFormer~\cite{song2022tempformer} and  VRT~\cite{liang2024vrt} is provided in \colorbox{gray!15}{\textcolor{MyDarkRed}{Supp. 3}}. Furthermore, we modify DPU by concatenating temporal frames along the channel dimension, denoted as DPU$^\ast$. Given the computational resource constraints, we set 5 stages as the DUNs benchmark.

\textbf{Implementation Details.} PG-SVRT is implemented in PyTorch  with the number of modules \( (N_1, N_2, N_3) = (4, 8, 8) \).  We set the basic channel size \( C = N_\lambda = 30 \) and frame number \( T = 3 \).  Training is conducted on RTX 3090 GPUs for 80 epochs with a batch size of 2. The multi-stage root mean square error (RMSE) loss function is used with the Adam optimizer, set with \( \beta_1 = 0.9 \), \( \beta_2 = 0.999 \). The initial learning rate is set to \( 3 \times 10^{-4} \), and a cosine annealing schedule is employed, with the learning rate gradually reduced to \( 1 \times 10^{-6} \) over the course of training.

\textbf{Evaluation Metrics.} We use PSNR, SSIM~\cite{wang2004image}, and SAM~\cite{kruse1993spectral} to evaluate the image quality and spectral fidelity, while ST-RRED~\cite{soundararajan2012video} measures temporal consistency. FLOPs and Params are used to assess model complexity. All metrics, except for Params, are reported as frame-wise averages. 

\subsection{Evaluation on Representative SCI Systems}
\label{sec_exp_system}
\vspace{-12pt}

\input{figure_tex/table1}
\vspace{-18pt}
\input{figure_tex/figure5}

Given the distinct advantages and limitations of each system, we conduct a comparative simulation study across four representative SCI architectures (SD-CASSI~\cite{wagadarikar2008single}, DD-CASSI~\cite{gehm2007single}, PMVIS~\cite{cao2011prism}, and NDSSI~\cite{chen2023notch}) to assess their potential in spectral representation and spatiotemporal feature extraction. As shown in Tab.~\ref{tab_architecture} and Fig.~\ref{fig_measurements}, under our tested configurations, DD-CASSI achieves the highest reconstruction quality. While PMVIS and SD-CASSI lack the spatial cues necessary for temporal propagation, and NDSSI’s sparse sampling limits spectral capacity, DD-CASSI benefits from clear structural representation and efficient encoding, which allow it to excel in video-level spectral reconstruction tasks. To address potential generalization concerns, we provide broader stress tests under varying noise and spectral settings in \colorbox{gray!15}{\textcolor{MyDarkRed}{Supp. 2}}, which further confirm DD-CASSI's robustness within its expected operating regime. Consequently, we adopt DD-CASSI as the base system for subsequent evaluations and prototype construction.

\subsection{Quantitative Comparisons with SOTA Methods}

As shown in Tab.~\ref{tab_simexp},  PG-SVRT achieves superior results in spatial quality, spectral fidelity, and temporal consistency.  Although DPU$^\ast$ leverages temporal information to improve certain metrics, it requires substantially higher computational cost. In contrast, PG-SVRT, despite being a video-based model, achieves lower per-frame FLOPs than several image-based methods, indicating its efficiency in modeling spatiotemporal dependencies.  As indicated by the SAM comparison,  most methods exhibit high uncertainty when reconstructing masked spectral signals, whereas PG-SVRT leverages complementary information across adjacent frames to compensate for missing spectral details, thereby improving   spectral fidelity.  Moreover, the ST-RRED scores suggest that our method maintains stronger temporal consistency, laying a reliable foundation for  spectral video perception.

\input{figure_tex/table2}

\subsection{Qualitative Analysis}

\textbf{Simulation.}  Fig.~\ref{fig_simexp} presents the visual results of PG-SVRT and comparison methods. While all methods appear to restore spatial details across spectral channels, benefiting from the clear structural information provided by DD-CASSI, the error maps (bottom-right of each subfigure) show that PG-SVRT exhibits minimal errors, indicating superior spectral fidelity. Furthermore, spectral curves show that PG-SVRT achieves a higher correlation with the ground truth, matching the shapes more closely.

\input{figure_tex/figure6}
\input{figure_tex/figure7}
\input{figure_tex/figure8}

\vspace*{-4pt}
\textbf{Real Experiments.} 
We further evaluate PG-SVRT in real-world scenarios. All methods are retrained on DynaSpec using the real mask. As shown in Fig.~\ref{fig_realexp}, we synthesize pseudo-RGB images from the reconstructed HSIs for comprehensive visual assessment. Notably, the Winnie-the-Pooh head reconstructed by comparison methods exhibits distortions and striping, while PG-SVRT yields more natural results. To further validate robustness, we challenge PG-SVRT with scenes containing more complex textures, as illustrated in Fig.~\ref{fig_realpre}. Both the pseudo-RGB and spectral images demonstrate the reliability and practicality of PG-SVRT under real-world conditions. Additionally, more reconstruction comparison results for real measurements are provided in \colorbox{gray!15}{\textcolor{MyDarkRed}{Supp. 4}}.

\section{Ablation Study.} 

\textbf{Break-down Ablation.}  We adopt a baseline,  comprising  a spatially-windowed F-MSA~\cite{bertasius2021space} and regular FFN of MST~\cite{cai2022mask}, to study the effect of each principal component, as shown in  Tab.~\ref{tab3_breakdown}.   Taking PSNR as an example, the baseline yields 39.97~dB, while PG-SVRT achieves improvements of 1.33 dB, 0.11 dB, and 0.11 dB when CDPA, MGDP, and MDFFN are successively applied. These results validate the effectiveness of proposed modules. In particular, CDPA plays a key role by leveraging inter-frame complementary features to enhance reconstruction quality and temporal consistency. Additionally, a comparison of different attention mechanisms is provided in the \colorbox{gray!15}{\textcolor{MyDarkRed}{Supp. 5}}.

\input{figure_tex/table3_4}
\vspace*{-4pt}

\textbf{Multi-domain Information Process.}  We study different spatiotemporal signal processing strategies, as shown in Tab.~\ref{tab4_multidomain}. The strategies are categorized into three main types: parallel processing (Parallel), temporal-then-spatial (T-S), and spatial-then-temporal (S-T). Additionally, $w/~P$ denotes our value-sharing propagation mechanism.  In the horizontal comparison without the propagation strategy, the parallel scheme focuses more on intra-domain features, effectively reducing cross-domain interference and thereby improving reconstruction performance. Similarly, in our propagation mechanism, the $Q$ and $K$ are generated prior to any feature processing, which ensures that $QK$ interactions are not affected by crosstalk. Moreover, the value-sharing strategy promotes inter-domain feature fusion, ultimately contributing to enhanced spectral reconstruction performance.

\input{figure_tex/table5_6}

\textbf{Bridged Token Ablation.} As shown in Tab.~\ref{tab5_bridgedtoken},  we analyze the impact of introducing Bridged Tokens ($B_S$). \textit{None} indicates the absence of $B_S$,  while other values represent the number of tokens \( N_B \). When the condition \( 2N_B < H_{win} W_{win} \) is satisfied (e.g., \( N_B = 16 \) or \( 64 \)), the model achieves lower computational complexity while maintaining comparable or even superior reconstruction performance. Otherwise (e.g., $N_B = 144$), performance still improves, but at the cost of significantly higher  complexity. Conceptually, $B_S$ can be regarded as representative embeddings of $Q$. By establishing associations between them, $B_S$ extracts core  information from \( Q \) and interacts with \( K \) and \( V \), enabling efficient feature interaction while reducing computational overhead.

\textbf{Improvements of MDFFN.} We conduct an ablation study on MDFFN, as shown in Tab.~\ref{tab6_ab_mdffn}. While regular conv3d-based FFN has large receptive fields, they struggle to extract useful information from redundant features. Single-domain processing ignores temporal or spatial continuity priors. In contrast, applying intra-domain self-attention followed by cross-domain fusion enables more effective feature extraction and leads to improved reconstruction quality.

%% file: figure_tex/table1.tex

\begin{table}[H]
\caption{Quantitative comparison of representative SCI architectures. All systems are solved via PG-SVRT under a unified mathematical framework to exclude algorithmic bias.}
\label{tab_architecture}
            \vspace{-0.3em}
            \centering
            \renewcommand{\arraystretch}{1.05} 
            \resizebox{1\linewidth}{!}{ 
                \begin{tabular}{ccccc}
                    \hline
                    & PMVIS  & SD-CASSI & NDSSI  & DD-CASSI \\ \hline
                    PSNR$~\uparrow$    & 28.45  & 37.78    & {\second{37.84}}  & \best{41.52}    \\
                    SSIM$~\uparrow$    & 0.8456 & 0.9700   & {\second{0.9825}} & \best{0.9893}   \\
                    SAM$~\downarrow$     & 5.4162 & {\second{4.0737}}   & 5.4091 & \best{3.9084}   \\
                    ST-RRED$~\downarrow$ & 459.49   & \best{23.21}     & 91.8   & {\second{23.25}}    \\ \hline
                \end{tabular}
            }
\end{table}

%% file: figure_tex/figure5.tex
\begin{figure}[H]
    \centering
    \includegraphics[width=1\linewidth]{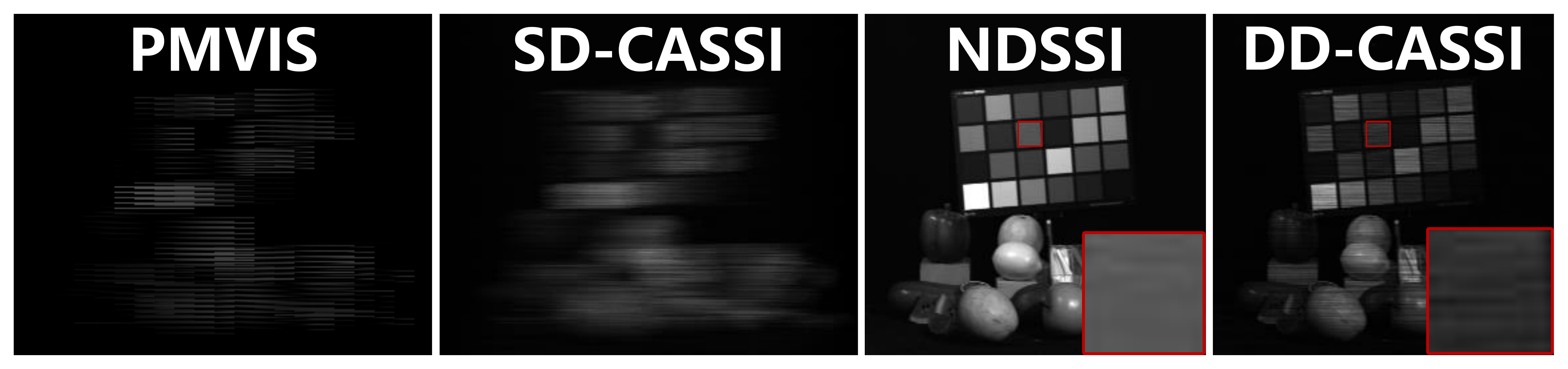} 
    \vspace{-14pt}
    \caption{Measurements of different SCI systems}
    \vspace{-12pt}
    \label{fig_measurements}
\end{figure}

%% file: figure_tex/table2.tex
\begin{table*}[t!]
\caption{Quantitative comparisons of several SOTA methods and PG-SVRT. The suffix -K denotes results on the KAIST, while -D represents evaluations on the DynaSpec testset. The \textbf{best} and \underline{second best} results are highlighted in bold and underline, respectively. \label{tab_simexp}}
\centering
\renewcommand{\arraystretch}{1.1}
\setlength{\tabcolsep}{2pt} 
\resizebox{0.97\textwidth}{!}{
\begin{tabular}{ccccccccccccc}
\hline
Method    & \makecell[c]{MST-L \\[-4pt] \scriptsize CVPR 2022} 
          & \makecell[c]{CST-L \\[-4pt] \scriptsize ECCV 2022} 
          & \makecell[c]{DAUHST \\[-4pt] \scriptsize NeurIPS 2022} 
          & \makecell[c]{GAP-Net \\[-4pt] \scriptsize IJCV 2023} 
          & \makecell[c]{\footnotesize DADF-Plus-3 \\[-4pt] \scriptsize TMI 2023} 
          & \makecell[c]{RDLUF \\[-4pt] \scriptsize CVPR 2023} 
          & \makecell[c]{PADUT \\[-4pt] \scriptsize ICCV 2023} 
          & \makecell[c]{ $S^2$-Transfor. \\[-4pt] \scriptsize TPAMI 2024} 
          & \makecell[c]{SSR \\[-4pt] \scriptsize CVPR 2024} 
          & \makecell[c]{DPU \\[-4pt] \scriptsize CVPR 2024} 
          & \makecell[c]{DPU$^\ast$ \\[-4pt] \scriptsize CVPR 2024} 
          & \makecell[c]{PG-SVRT \\[-4pt] Ours} \\ \hline
PSNR-K$~\uparrow$    & 39.99                                  & 39.93     & 38.98        & 36.92     & 38.23       & 39.26     & 38.61     & 33.26                                  & 39.04     & 40.02                                  & {\second{40.50}} & {\best{41.23}}  \\
SSIM-K$~\uparrow$    & {\second{0.9881}} & 0.9864    & 0.9832       & 0.9755    & 0.9832      & 0.9860    & 0.9828    & 0.9617                                 & 0.9842    & 0.9856                                 & 0.9853                                & {\best{0.9882}} \\
SAM-K$~\downarrow$     & {\second{3.8248}} & 4.1342    & 5.4514       & 6.1204    & 4.7676      & 4.2932    & 4.7154    & 8.0837                                 & 5.2201    & 5.2250                                 & 5.1685                                & {\best{3.805}}  \\
ST-RRED-K$~\downarrow$ & 30.99                                  & 35.11     & 37.27        & 85.34     & 48.45       & 39.06     & 47.19     & 155.82                                 & 38.29     & {\second{25.90}}  & 26.71                                 & {\best{19.35}}  \\ \hline
PSNR-D$~\uparrow$    & 39.58                                  & 40.06     & 40.39        & 39.38     & 39.00       & 39.26     & 40.41     & 37.10                                  & 39.66     & 41.01                                  & {\second{41.36}} & {\best{41.82}}  \\
SSIM-D$~\uparrow$    & 0.9873                                 & 0.9876    & 0.9883       & 0.9851    & 0.9861      & 0.9863    & 0.9881    & 0.9786                                 & 0.9873    & {\second{0.9893}} & 0.9889                                & {\best{0.9904}} \\
SAM-D$~\downarrow$     & {\second{4.2208}} & 4.4578    & 4.7962       & 5.3402    & 4.6057      & 4.4429    & 4.4372    & 6.0231                                 & 4.6840    & 4.4732                                 & 4.5997                                & {\best{4.0118}} \\
ST-RRED-D$~\downarrow$ & 66.31                                  & 52.19     & 46.64        & 67.54     & 73.17       & 70.64     & 48.88     & 114.17                                 & 59.03     & 36.84                                  & {\second{31.20}} & {\best{27.14}}  \\ \hline
Params    & 2.31 M                                 & 3.44 M    & 3.36 M       & 4.28 M    & 20.25 M     & 2.17 M    & 2.57 M    & {\best{1.33 M}} & 2.06 M    & {\second{1.88 M}} & 15.14 M                               & 2.48 M                                 \\
GFLOPs    & {\second{28.23}}  & 28.53     & 35.93        & 58.15     & 76.33       & 59.69     & 32.78     & 56.17                                  & 29.92     & 31.04                                  & 77.36                                 & {\best{28.18}}  \\ \hline
\end{tabular}

}
\vspace*{-8pt}
\end{table*}

%% file: figure_tex/figure6.tex
\begin{figure*}[t!]
\centering
\includegraphics[width=0.95\textwidth]{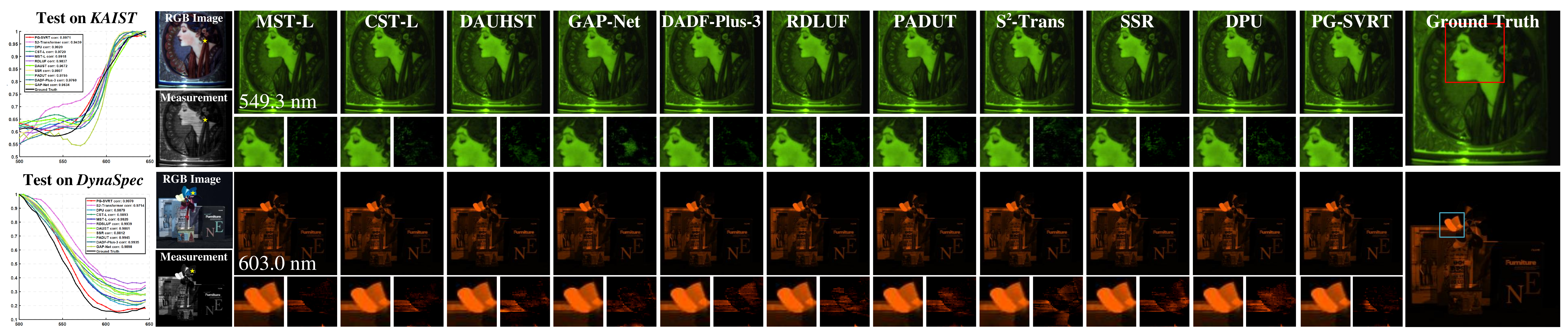}
\vspace*{-10pt}
\caption{Reconstruction results of PG-SVRT and comparison methods on the KAIST and DynaSpec test sets. The bottom-left corner of each subplot presents an enlarged detail view, while the bottom-right corner shows the difference with the GT. It is evident that, while all methods benefit from DD-CASSI and are able to recover structural details, our method achieves the superior fidelity.}
\label{fig_simexp}
\vspace*{-5pt}
\end{figure*}

%% file: figure_tex/figure7.tex
\begin{figure*}[t!]
\centering
\includegraphics[width=0.95\textwidth]{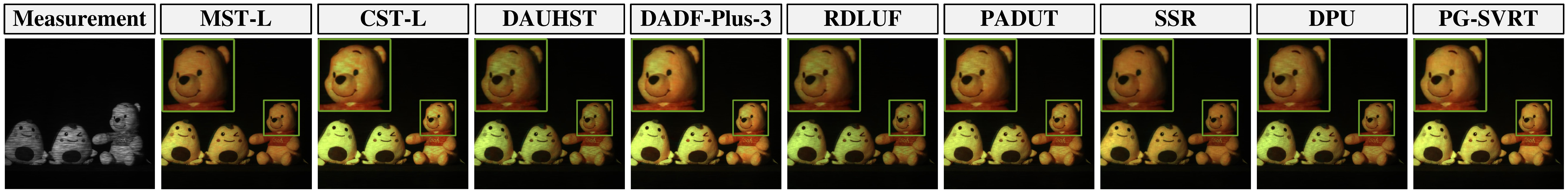}
\vspace*{-8pt}
\caption{Reconstruction of real measurements using comparison methods and PG-SVRT, with  pseudo-RGB images generated from the reconstructed HSIs to assess reconstruction quality across all bands. Compared to other methods, PG-SVRT results exhibit fewer artifacts.}
\label{fig_realexp}
\vspace*{-5pt}
\end{figure*}

%% file: figure_tex/figure8.tex
\begin{figure*}[t!]
\centering
\includegraphics[width=0.95\textwidth]{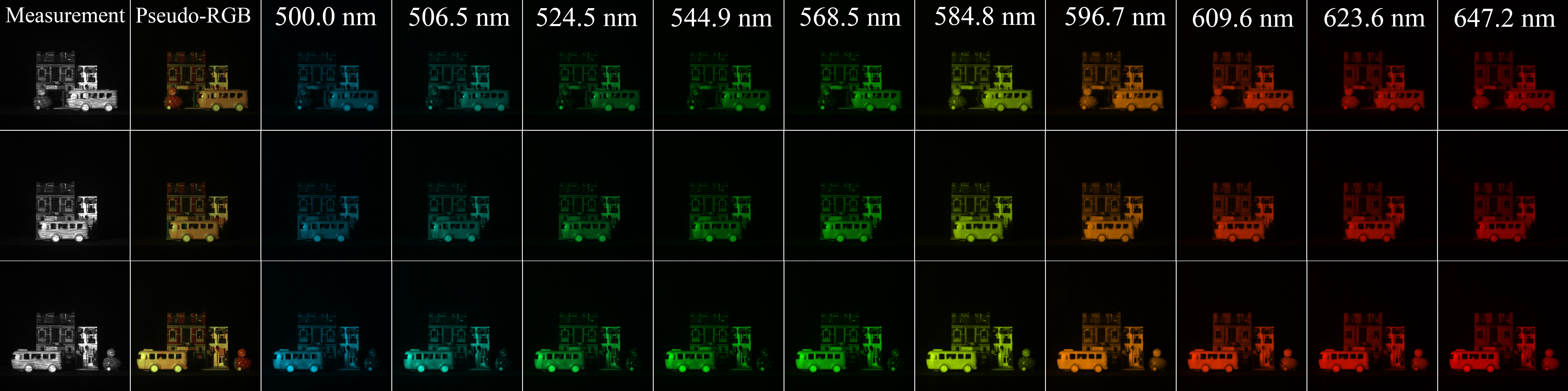}
\vspace*{-8pt}
\caption{Complex real-world scenes reconstructed by PG-SVRT show clear structures and fine details in both the pseudo-RGB and HSIs.}
\label{fig_realpre}
\vspace*{-18pt}
\end{figure*}

%% file: figure_tex/table3_4.tex

\begin{table}[H]
            \caption{Break-down ablation study of PG-SVRT,  "+" indicates adding or replacing modules relative to the baseline.}
            \label{tab3_breakdown}
            \centering
            \small
            \renewcommand{\arraystretch}{1.0} 
            \setlength{\tabcolsep}{3pt}       
                \begin{tabular}{ccccc}
\hline
       & Baseline & + CDPA & + MGDP & + MDFFN \\ \hline
PSNR   & 39.97    & 41.30  & 41.41  & {\best{41.52}}   \\
SSIM   & 0.9827   & 0.9884 & 0.9886 & {\best{0.9893}}  \\
SAM    & 5.5312   & 4.3224 & 4.2513 & {\best{3.9084}}  \\
STRRED & 43.90    & 25.44  & 24.63  & {\best{23.25}}   \\ \hline
Params & 2.17 M & 1.92 M & 1.92 M & 2.48M\\
GFLOPs & 30.11 & 21.11 & 21.31 & 28.18\\ \hline
\end{tabular}
        \end{table}
\vspace{-12pt} 
\begin{table}[H]
            \caption{Comparison of spatiotemporal processing strategies.}
            \label{tab4_multidomain}
            \centering
            \renewcommand{\arraystretch}{1.0} 
            \setlength{\tabcolsep}{3pt}       
            \small
\begin{tabular}{cccccc}
\hline
       & Parallel & T-S    & T-S w/~P & S-T    & S-T w/~P                                 \\ \hline
PSNR   & 41.35    & 41.04  & 41.47   & 41.08  & {\best{41.52}}  \\
SSIM   & 0.9886   & 0.9877 & 0.9892  & 0.9880 & {\best{0.9893}} \\
SAM    & 4.1518   & 4.4995 & 3.9605  & 4.4128 & {\best{3.9084}} \\
STRRED & 26.00    & 28.67  & 26.60   & 26.40  & {\best{23.25}}  \\ \hline
Params & 2.85 M & 2.60 M & 2.48 M & 2.60 M & 2.48 M \\
GFLOPs & 33.23 & 30.16 & 28.18 & 30.16 & 28.18 \\ \hline
\end{tabular}
        \end{table}

%% file: figure_tex/table5_6.tex
\begin{figure*}[t!]
    \begin{minipage}{0.59\textwidth}
        \centering
        \begin{table}[H]
            \caption{Ablation on the number of bridged tokens.}
            \label{tab5_bridgedtoken}
            \vspace{-0.3em}
            \centering
            \renewcommand{\arraystretch}{1.05}
            \setlength{\tabcolsep}{3pt}
            \resizebox{1\textwidth}{!}{ 
\begin{tabular}{cccccccc}
\hline
$N_B$   & \textless{}~$H_{win}W_{win}/2$ & PSNR                                  & SSIM                                   & SAM                                    & ST-RRED                               & GFLOPs                                & Params                                 \\ \hline
None & \textbf{--}                   & 41.18                                 & 0.9883                                 & 4.1254                                 & 27.90                                  & 32.20                                  & {\best{2.36 M}} \\
16   & \ding{51}                   & 41.11                                 & 0.9882                                 & 4.1359                                 & 28.89                                 & {\best{25.26}} & 2.37 M                                 \\
64   & \ding{51}                    & {\best{41.52}} & {\best{0.9893}} & {\best{3.9084}} & {\best{23.25}} & 28.18                                 & 2.48 M                                 \\
144  & \ding{55}                   & 41.34                                 & 0.9887                                 & 4.0881                                 & 26.15                                 & 33.10                                  & 2.74 M                                 \\ \hline
\end{tabular}
            }
        \end{table}
    \end{minipage}%
    \hfill
    \begin{minipage}{0.38\textwidth}
        \centering
        \begin{table}[H]
            \caption{Ablation study on MDFFN.}
            \label{tab6_ab_mdffn}
            \vspace{-0.6em}
            \centering
            \renewcommand{\arraystretch}{1.05}
            \setlength{\tabcolsep}{2pt}
            \resizebox{1\textwidth}{!}{ 
\begin{tabular}{ccccc}
\hline
               & PSNR                                  & SSIM                                   & SAM                                    & ST-RRED                               \\ \hline
regular Conv3d & 41.41                                 & 0.9886                                 & 4.2513                                 & 24.63                                 \\
w/o temporal         & 41.16                                 & 0.9882                                 & 4.1203                                 & 26.71                                 \\
w/o spatial         & 41.42                                 & 0.9885                                 & 4.123                                  & 26.18                                 \\
MDFFN          & {\best{41.52}} & {\best{0.9893}} & {\best{3.9084}} & {\best{23.25}} \\ \hline
\end{tabular}
            }
        \end{table}
    \end{minipage}
    \vspace*{-16pt}
\end{figure*}

%% file: sec/6_conclusion.tex
\vspace{-10pt}
\section{Conclusion}
\vspace{-6pt}

In this work, we tackle the challenges of video-level compressive spectral reconstruction by constructing the DynaSpec dataset, DD-CASSI prototype, and PG-SVRT Network, which effectively fuses spatiotemporal features with low computational complexity.  While DynaSpec serves as a valuable controlled benchmark, we recognize that its idealized acquisition settings (indoor lighting and specific motion statistics) may limit generalization to unscripted, unknown natural environments. Nevertheless, extensive experiments demonstrate that our method achieves superior performance and robust trends under domain shifts. Collectively, it represents a foundational step toward high-dimensional spectral video reconstruction, catalyzing further research into more complex, open-world applications.